\documentclass[10pt,logo,copyright]{nvidiatechreport}
\usepackage[numbers,sort]{natbib}
\usepackage[capitalize]{cleveref}
\crefname{algocf}{Algorithm}{Algorithms}
\Crefname{algocf}{Algorithm}{Algorithms}

\usepackage[ruled,linesnumbered]{algorithm2e}

\SetAlFnt{\small}
\SetAlCapFnt{\small}
\SetAlCapNameFnt{\small}
\SetAlCapHSkip{0pt}

\SetCommentSty{smallitcomment}

\title{HiGS: A Hierarchical Rendering Architecture for Real-Time 3D Gaussian Splatting}

\author{
  \textbf{Dawid Paj\k{a}k} \quad
  \textbf{Martin Bisson} \quad
  \textbf{Rodolfo Lima} \\
  \small NVIDIA \\  
}

\begin{document}
\maketitle

\begin{figure}[!ht]
\includegraphics[width=\textwidth]{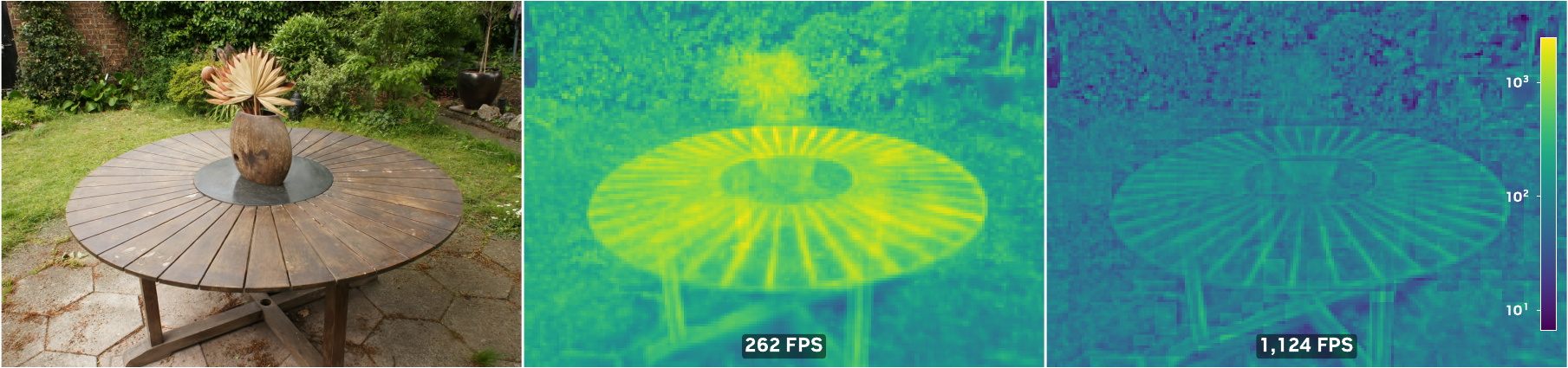}
\caption{\textbf{HiGS} renders a 5.8M-gaussian scene at 1{,}124~FPS (1080p). \emph{Left:} rendered image. \emph{Center} and \emph{Right} show how evenly rendering work is spread, measured per tile as the gaussians covering the tile divided by the number of parallel work units that work on it. \emph{Center:} gsplat~\cite{ye2024gsplat} uses one work unit per $8\times8$ tile, so the value is the raw per-tile gaussian count and dense regions form severe hotspots (peak 2{,}485). \emph{Right:} HiGS groups tiles into $8\times4$ macro-tiles and splits each macro-tile's gaussians into \emph{independent} batches of at most 1{,}024.
Dense macro-tiles spawn proportionally more batches, so the per-work-unit load flattens and its peak across tiles drops to ${\sim}$340.}
\label{fig:teaser}
\end{figure}

\begin{abstract}
3D Gaussian Splatting (3DGS) has become the standard for real-time novel view synthesis on commodity GPUs.
Its pipeline ties spatial partitioning and rasterization to one tile size, yet the two pull in opposite directions: partitioning, which bins and depth-sorts gaussians, grows cheaper with larger tiles, while rasterization gets cheaper with smaller ones.
Prior acceleration work reduces the cost of individual stages but keeps both locked to that single scale, where a few dense tiles dominate frame time.
We present \emph{Hierarchically Tiled Gaussian Splatting} (HiGS), which gives each its own scale: partitioning runs over coarse \emph{macro-tiles}, while rasterization runs over the fine render tiles within them.
Rasterization work is then issued in proportion to the gaussians in each macro-tile rather than per tile, so dense regions spread across many parallel units instead of serializing through one.
Across tested scenes, HiGS renders up to {\boldmath${\sim}15.8\times$} faster than the original 3DGS and outperforms every other rasterizer we evaluate, while preserving exact front-to-back alpha compositing.

\end{abstract}

\abscontent

\section{Introduction}
\label{sec:intro}

Real-time photorealistic rendering of three-dimensional scenes reconstructed from multi-view captures has become foundational to virtual reality, telepresence, and immersive content creation.
3D Gaussian Splatting~\cite{kerbl2023gaussian} has emerged as the leading approach, combining the visual fidelity of neural radiance fields~\cite{mildenhall2020nerf} with the throughput of an explicit, differentiable representation that maps efficiently to commodity GPU rasterization.
Its open-source implementation gsplat~\cite{ye2024gsplat} has become the de facto foundation for subsequent work on training, compression, and rendering.

Most variants of 3DGS share a similar pipeline factorization: project gaussians to screen space, assign each to overlapping render tiles, globally sort all (tile, depth) pairs, then rasterize each tile independently.
This factorization maps cleanly onto GPU rasterization primitives, but also introduces a fundamental trade-off at its core related to the tile size.
Partitioning favors large tiles: a larger spatial cell is less likely to be straddled by any single gaussian's screen-space extent, so the (tile, gaussian) pair count---and every dependent stage of binning, sort, and intersection testing---shrinks with tile area.
Rasterization, on the other hand, favors small tiles.
As tile area grows, more gaussians overlap each tile and a larger fraction of pixel--gaussian evaluations fall outside the gaussian's effective support, inflating per-pixel work with computations that contribute nothing to the final image.
The one-block-per-tile dispatch compounds the imbalance: a few dense tiles dominate frame time.
3DGS ties both stages to a single render-tile size and pays both costs: the global radix sort over composite tile-depth keys processes tens of millions of pairs per frame at 4K and consumes several milliseconds, while rasterizer time varies by several multiples across viewpoints of the same scene (\cref{fig:work-decomp}, left).

A growing body of work targets these costs along two axes: input-side reduction through pair compaction, pruning, and model compression~\cite{hanson2024speedysplat,feng2024flashgs,wang2024adrgaussian,radl2024stopthepop,hahlbohm2026fastergs,fan2024lightgaussian}, and in-pipeline acceleration through kernel-level optimization, restructured rasterizer parallelism, or two-pass sort variants~\cite{liao2025tcgs,schutz2025splatshop,gui2024balanced}.
These approaches retain the shared-tile-size constraint and the trade-off it forces.
A few software approaches do depart from this factorization, but pay for it with approximate compositing or alternative rendering substrates~\cite{hou2025sortfree,rueckert2025stochastic,yuan2025hardware}.
A separate hardware line resolves the trade-off architecturally---by sorting at coarser-than-tile granularity while preserving exact compositing---but realizes the design on custom hardware datapaths~\cite{lee2024gscore,jo2025gstg}.

We present HiGS, a rendering architecture that resolves the same trade-off on commodity GPUs by decoupling partitioning from rasterization granularity.
Gaussians are binned and depth-sorted at a coarse \emph{macro-tile} granularity---a grouping of render tiles processed jointly---while a rasterization kernel dispatches at the finer render-tile granularity, determining per-tile visibility inline and reusing each loaded gaussian across all member tiles from local storage.
The work decomposition transitions from \emph{(tile)} to \emph{(macro-tile, gaussian-batch)}: dense macro-tiles spawn proportionally more bounded-work units, so resources scale with gaussian density rather than screen area, eliminating the tail effect by construction.
From this single decoupling, segmented sorting, cross-tile data reuse, and density-proportional load balancing emerge as architectural consequences.
Across every tested scene and resolution, HiGS is faster than every rasterizer in our benchmark, with no observable image-quality loss.

Our contributions form a single pipeline built around the \emph{macro-tile} primitive:
\begin{itemize}
    \item a \textbf{two-scale spatial hierarchy} that decouples partitioning from rasterization granularity, resolving the tile-size trade-off at 3DGS's core;
    \item a \textbf{dynamically load-balanced rasterizer} that computes per-tile visibility inline during the cooperative gaussian load and balances work at three levels---dense macro-tiles spawn more processing units, worker groups claim render tiles from a shared queue, and a per-worker scheduling buffer absorbs per-tile density variation before fixed-rate blending---eliminating the rasterizer tail effect on commodity GPUs;
    \item a \textbf{fully asynchronous on-device segmented depth sort} that runs as a cascade of tier-classified per-segment block sorts, with oversized segments handled by a device-built merge tree consumed by a persistent kernel, scaling across orders of magnitude of segment sizes without host involvement;
    \item a \textbf{half-precision data path} stabilized by Cholesky factors of conics, which turn the Mahalanobis quadratic into a sum of squares free of fp16 cancellation.
\end{itemize}

\section{Related Work}
\label{sec:related}

Kerbl~\etal~\cite{kerbl2023gaussian} established the tile-based rendering architecture for 3D Gaussian Splatting: project gaussians to 2D, assign each to overlapping tiles via axis-aligned bounding boxes, globally sort all (tile, depth) pairs, and rasterize each tile with one thread block.
gsplat~\cite{ye2024gsplat} popularized this design through a widely adopted open-source implementation.
Subsequent work falls along three axes---reducing the input to the pipeline, accelerating the pipeline itself, and improving rendering fidelity or training efficiency.

\noindent\textbf{Pair reduction.}
A line of research targets the false-positive tile--gaussian pairs that drive sort and rasterization cost.
Opacity-aware bounding---computing the ellipse extent at the alpha detection threshold rather than at a fixed number of standard deviations---was introduced by StopThePop~\cite{radl2024stopthepop}, FlashGS~\cite{feng2024flashgs}, and AdR-Gaussian~\cite{wang2024adrgaussian}; Speedy-Splat~\cite{hanson2024speedysplat} and Faster-GS~\cite{hahlbohm2026fastergs} adopt and extend the technique.
Tighter intersection tests followed: Speedy-Splat's AccuTile traces per-row ellipse boundaries, AdR-Gaussian derives per-axis rectangular bounds from diagonal covariance entries, and FlashGS performs exact ellipse--rectangle intersection via algebraic edge comparisons.
FlashGS additionally combines these reductions with kernel-level micro-optimizations across the projection and rasterization stages.
Like these methods, HiGS reduces pair counts; in contrast, it does so by partitioning at coarser-than-render-tile granularity rather than by refining the per-tile assignment.

\noindent\textbf{Model-side reduction.}
Model-side reduction complements pair reduction by shrinking the gaussian set itself.
Speedy-Splat applies sensitivity-based pruning during training, LightGaussian~\cite{fan2024lightgaussian} performs vector-quantized model compression at training time, RadSplat~\cite{niemeyer2024radsplat} combines NeRF-supervised pruning with test-time visibility filtering to reduce the gaussian set by up to $10\times$, and Taming~3DGS~\cite{mallick2024taming} replaces 3DGS's gradient-driven densification with a score-guided procedure under a user-specified budget, producing $4$--$5\times$ smaller models at comparable quality.
A parallel line of learned-codebook quantization approaches shrinks per-gaussian state: EAGLES~\cite{girish2024eagles} represents color, rotation, and opacity as MLP-decoded integer latents, Compact-3DGS~\cite{lee2024compact3dgs} combines a learnable volume mask with R-VQ geometry codebooks and a hash-grid color field replacing per-gaussian spherical harmonics, and Compressed 3DGS~\cite{niedermayr2024compressed} applies sensitivity-aware k-means clustering with quantization-aware fine-tuning and entropy coding.
Mini-Splatting~\cite{fang2024minisplatting} reorganizes the gaussian distribution itself via blur split, depth reinitialization, and importance-weighted sampling; RetinaGS~\cite{li2024retinags} scales 3DGS to billion-gaussian scenes via multi-GPU model parallelism.
All of these input-side reductions compose with renderer-side acceleration: a compacted or compressed gaussian set fed to HiGS would compound the input reduction with the architectural speedup.

\noindent\textbf{Rendering acceleration.}
A second line of work accelerates the rendering pipeline itself.
Beyond pair-count reduction, FlashGS fuses the preprocessing kernels, applies software pipelining across the dependent gaussian-fetch chain, and uses base-2 exponentials evaluated through the GPU's special function unit.
Splatshop~\cite{schutz2025splatshop} narrows sort keys by sorting visible splats by 32-bit depth before tile-fragment creation, reducing total sort cost by $1.5$--$1.9\times$.
TC-GS~\cite{liao2025tcgs} reformulates the alpha-blending exponent as a Tensor-Core matrix multiplication, reporting up to $2.1\times$ speedup over 3DGS and further gains when layered on prior accelerators.
Balanced 3DGS~\cite{gui2024balanced} addresses the rasterizer's tail effect through dynamic inter-block tile dispatch, gaussian-wise warp collaboration with prefix-product transmittance shuffles, and fine-grained 4-pixel scheduling units.
A separate group of approaches departs from exact front-to-back compositing: Sort-free Gaussian Splatting~\cite{hou2025sortfree} replaces alpha blending with a commutative weighted sum, removing the depth-ordering requirement and per-pixel early termination while reporting PSNR comparable to alpha-blended 3DGS, while StochasticSplats~\cite{rueckert2025stochastic} uses stochastic transparency over standard Z-buffering, exposing a samples-per-pixel knob that trades latency for image quality.
Closer to the metal, Yuan and He~\cite{yuan2025hardware} port the full pipeline to Vulkan---hardware blending for forward, programmable blending via fragment-shader interlock for backward---reaching $3\times$ forward speedup at fp16, at the cost of per-pixel transmittance early-termination and software-managed shared memory.
Hardware proposals go further still: GSCore~\cite{lee2024gscore} proposes a custom ASIC combining shape-aware oriented-bounding-box intersection, hierarchical sorting that overlaps with rasterization, and a per-subtile visibility bitmap, GauRast~\cite{li2025gaurast} extends the GPU triangle rasterizer at the RTL level with native gaussian alpha evaluation, and GSAcc~\cite{yang2025gsacc} proposes a custom ASIC that exploits temporal depth speculation across frames combined with a gaussian-centric dataflow.
Jo and Park~\cite{jo2025gstg} propose GS-TG, an ASIC accelerator that decouples sort and rasterization granularities.
Of these, GS-TG is the closest in spirit to HiGS, sharing the same two-level decoupling of coarse sorting from fine rasterization.
The authors report, however, that on a GPU its per-gaussian bitmask cannot be generated concurrently with the sort, leaving preprocessing slower than the baseline and motivating their dedicated accelerator.
Unlike these acceleration approaches, which fuse stages, swap compositing models, or move to alternative hardware, HiGS restructures the partitioning--rasterization hierarchy while retaining the standard exact-alpha rendering model on a commodity GPU.

\begin{figure*}[t]
\centering
\includegraphics[width=\textwidth]{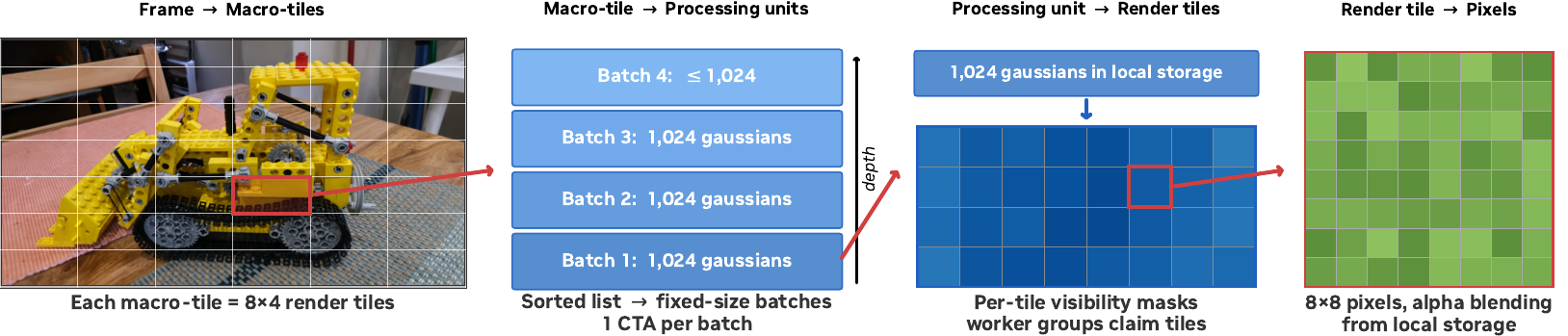}
\caption{\textbf{Hierarchical work decomposition.} HiGS organizes rendering coarse-to-fine, from macro-tile groupings down to per-pixel blending. Render tiles (8$\times$8 pixels) are grouped into macro-tiles of 8$\times$4 render tiles---the coarse unit at which binning and depth sort operate. Each macro-tile's sorted gaussian list is split into fixed-size batches of 1{,}024 gaussians, and each batch is assigned to one processing unit; dense macro-tiles spawn more processing units. Within each processing unit, per-tile visibility masks identify which gaussians overlap each render tile, and worker groups claim active tiles dynamically. Each worker group then rasterizes its 8$\times$8-pixel tile via front-to-back alpha blending, reading all data from local storage.}
\label{fig:pipeline}
\end{figure*}

\noindent\textbf{Quality, view consistency, and training.}
A complementary strand of work prioritizes rendering fidelity or training-time efficiency rather than primary-visibility throughput.
StopThePop~\cite{radl2024stopthepop} introduces per-ray depth ordering with a hierarchical rasterizer (warp-level, sub-tile, and per-pixel sort queues) and opacity-aware tile culling, eliminating popping artifacts at roughly $4\%$ overhead.
Mip-Splatting~\cite{yu2024mipsplatting} introduces a 3D smoothing filter that band-limits each primitive to its training-view Nyquist rate, paired with a 2D Mip filter approximating the physical pixel box filter, eliminating dilation and erosion artifacts under out-of-distribution sampling rates and adopted by several subsequent works as a default anti-aliasing front-end.
In~\cite{hahlbohm2025hybrid} authors replace the affine projection with perspective-correct Pl\"ucker-coordinate evaluation and combine a sorted alpha-compositing core with an order-independent tail accumulator.
3D Gaussian Ray Tracing \cite{moenneloccoz2024gaussianrt} departs from rasterization entirely via OptiX with icosahedral mesh proxies, enabling secondary ray effects at the cost of roughly $2\times$ slower primary visibility.
Complementing this, 3DGUT~\cite{wu20253dgut} retains rasterization while replacing the affine EWA projection with the unscented transform, propagating sigma points through the camera model to support distorted and rolling-shutter cameras; by aligning its formulation with that of 3D Gaussian Ray Tracing, the same representation supports a hybrid scheme that rasterizes primary rays and traces secondary rays for reflections and refractions.
On the training side, Taming~3DGS~\cite{mallick2024taming} introduces a per-splat backward parallelization that replaces 3DGS's per-pixel atomic-contention pattern with warp-collaborative gradient accumulation; Faster-GS~\cite{hahlbohm2026fastergs} integrates this scheme with a fused Adam optimizer and reports up to $5\times$ faster training end-to-end.
DISTWAR~\cite{durvasula2024distwar} addresses the same backward atomic-contention bottleneck through a complementary mechanism, a warp-level reduction primitive that aggregates gradient updates in registers before issuing one atomic per warp.
These directions are complementary to HiGS, which targets primary-visibility throughput at a fixed rendering model and integrates upstream of these refinements.

Throughput-focused prior work falls into two camps.
The pair-reduction, model-side, and rendering-acceleration approaches above retain the per-tile factorization: they reduce inputs, accelerate individual stages, or restructure rasterizer parallelism, but leave global sorting and per-tile scheduling intact.
The remaining approaches depart from that factorization at different costs: custom hardware datapaths (GSCore, GauRast, GSAcc, GS-TG) that sort locally and preserve exact compositing on dedicated silicon; compositing models that trade exact alpha blending or per-pixel early termination (Sort-free Gaussian Splatting, StochasticSplats, Yuan and He's Vulkan port); and ray tracing (3D Gaussian Ray Tracing), which preserves exact compositing at reduced primary-visibility throughput.
HiGS demonstrates that design directions recently explored for custom hardware pipelines translate efficiently to commodity GPUs. 
Realizing this required redesigning the pipeline around new partitioning, sorting and rasterization algorithms.

\section{Architecture}
\label{sec:arch}

HiGS is organized around a \emph{macro-tile hierarchy}: binning and depth sorting operate at coarse macro-tile granularity, while rasterization dispatches at the finer render-tile granularity within each macro-tile.
This reshapes work decomposition from screen-area-proportional to density-proportional and eliminates the rasterizer tail effect inherent to single-tile-size pipelines.

The remainder of this section walks the pipeline data flow shown in \cref{fig:pipeline}, from binning onward: the macro-tile hierarchy (\cref{sec:hierarchy}), gaussian partitioning (\cref{sec:binning}), segmented depth sort (\cref{sec:sort}), macro-tile rasterization with post-blend compositing (\cref{sec:fused-raster}), and the half-precision numerical representation (\cref{sec:numerical}).
The description is hardware-agnostic; GPU-specific mappings, including the standard fused projection and spherical-harmonic evaluation front-end~\cite{feng2024flashgs,hahlbohm2026fastergs}, are deferred to \cref{sec:impl}.

\subsection{Macro-tile hierarchy and work decomposition}
\label{sec:hierarchy}

Tile size in a tile-based gaussian splatting pipeline sits between two opposing pressures.
The partition-and-sort cost favors large tiles: a larger spatial cell is less likely to be straddled by any single screen-space gaussian extent, so the (tile, gaussian) pair count drops with tile area, shrinking the work for every binning, sort, and intersection-test stage downstream.
The rasterization cost favors small tiles: as tile area grows, more gaussians overlap each tile, each pixel pays more work, and a larger fraction of gaussian--pixel evaluations land outside the gaussian's effective support.
3DGS~\cite{kerbl2023gaussian} ties both stages to a single render-tile size and lives on one point along this trade-off; HiGS separates the two scales so each can be set independently.

The separation is realized as a two-level spatial hierarchy built bottom-up over the standard rasterization unit.
The \emph{render tile} ($8\times8$ pixels) is the fine-grained primitive shared with 3DGS-style pipelines; HiGS groups $M$ adjacent render tiles ($M{=}32$ in our implementation, arranged as $8\times4$ render tiles spanning $64\times32$ pixels) into a \emph{macro-tile}, the coarse-grained unit at which binning, depth sort, and data locality all operate.
Binning and depth sort run at macro-tile granularity (\cref{sec:binning,sec:sort})---segmented sort produces per-macro-tile depth-ordered gaussian lists---while rasterization dispatches at render-tile granularity (\cref{sec:fused-raster}), with per-render-tile visibility masks and per-render-tile blending.
Within each macro-tile, gaussian data is loaded into local storage once and reused across all $M$ member render tiles, amortizing the fetch over fine-grained output.

\begin{figure}[t]
\centering
\includegraphics[width=0.9\linewidth]{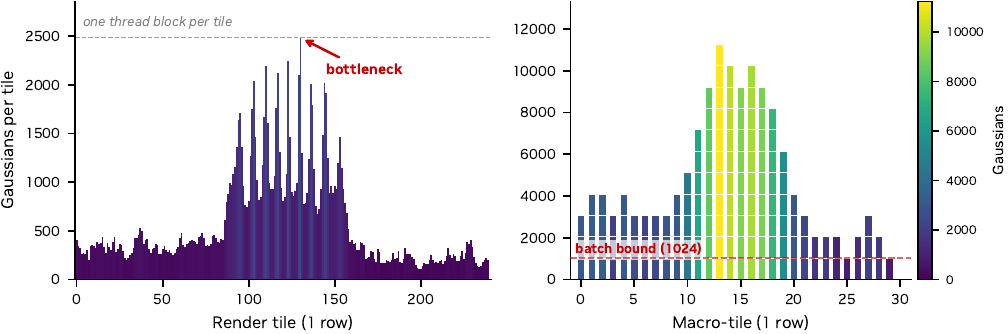}
\caption{\textbf{Work imbalance during rendering} (from \cref{fig:teaser} center scan-line). \emph{Left:} standard 3DGS assigns one thread block per tile which leads to load-imbalance and GPU under-utilization. \emph{Right:} HiGS groups tiles into coarser macro-tiles and divides each macro-tile's gaussians into fixed-size batches. Dense macro-tiles spawn more work that is both bounded and more concurrent at the thread block level.}
\label{fig:work-decomp}
\end{figure}

Work is decomposed coarse-to-fine to match.
Each macro-tile's depth-sorted gaussian list is split into fixed-size \emph{gaussian batches} of $B_r$ gaussians ($B_r{=}1{,}024$ in our implementation), and each (macro-tile, batch) pair becomes one independent \emph{processing unit}: a macro-tile holding $n$ gaussians spawns $\lceil n/B_r \rceil$ units.
Within each processing unit, render tiles are claimed dynamically by \emph{worker groups} from a shared \emph{tile work queue} $\mathcal{Q}$, and a per-worker \emph{scheduling buffer} $\mathcal{R}$ absorbs the per-tile variability in visible-gaussian count so the blending pipeline runs at a fixed rate.
The schedule is therefore density-proportional rather than screen-area-proportional: compute scales with gaussian count, and no single render tile can gate the frame the way the slowest tile gates a one-thread-group-per-tile assignment.
\Cref{fig:work-decomp} illustrates the decomposition.

The macro-tile decomposition trades memory for parallelism.
Per-macro-tile gaussian lists, per-batch partition histograms, and per-batch partial blending buffers are all preallocated up front so that binning batches, sort segments, and rasterization processing units run concurrently without dynamic allocation or producer--consumer stalls.
The resulting footprint scales with the number of macro-tile gaussian batches in the frame---roughly linearly in scene density at fixed resolution---and is an explicit consequence of replacing 3DGS's serial-by-tile model with a density-proportional schedule.

\subsection{Gaussian partitioning}
\label{sec:binning}

Binning constructs a per-macro-tile depth-keyed gaussian list.
Each macro-tile's gaussian count is data-dependent and unknown a priori; building per-macro-tile lists in parallel across millions of gaussians requires reconciling write positions without serializing through global atomics.
Kerbl~\etal~\cite{kerbl2023gaussian} address this by fusing list construction with depth ordering: a count pass records per-gaussian tile-overlap counts, a global prefix scan converts them to write offsets, and a fill pass emits composite-key records into an unordered global buffer; a subsequent radix sort over the composite key simultaneously groups records by tile and orders them by depth.
The per-tile list structure is materialized inside the sort, which is the structural reason that sort must be global.

Our binning produces per-macro-tile lists \emph{during} binning, leaving only in-segment ordering for the downstream sort.
Gaussians are processed in fixed-size partitioning batches of $B_p$ gaussians ($B_p{=}8{,}192$ in our implementation), each batch handled cooperatively by one work group: the work group maintains a private histogram in local storage with one counter per macro-tile in the rendered frame, and accumulates batch-scoped overlap counts via local atomics rather than global ones.
The accumulation fits in local storage because the frame's macro-tile count is modest---roughly 1{,}000 entries at 1080p and 4{,}000 at 4K with our $64\times32$-pixel macro-tile size.
A render-tile histogram at the same resolutions would carry 30{,}000--130{,}000 entries, exceeding commodity GPUs' local-storage capacity by an order of magnitude and forcing a global-atomic path.
Batch-local counts are then combined through a hierarchical prefix scan into per-macro-tile write offsets.
The fill pass replays the same batched decomposition and writes (depth, gauss-id) pairs directly into each gaussian's destination macro-tile range.
The output is the segmented structure consumed by \cref{sec:sort}: per-macro-tile lists of 32-bit depth keys ready for in-segment sorting, with no global composite-key sort required.

The gaussian-to-macro-tile intersection test uses a conservative per-row ellipse-contour scan applied at macro-tile granularity, following the AccuTile family~\cite{hanson2024speedysplat} but one level coarser.
Compared against a representative modern rasterizer---gsplat with opacity-aware AABB binning~\cite{ye2024gsplat,radl2024stopthepop}---mean pair counts drop from 10.1M to 1.51M at 1080p (85\% reduction) and 29.3M to 2.21M at 4K (92\%) (\cref{tab:pairs}).
The reduction splits between an AccuTile-style tile--ellipse intersection test and macro-tile coarsening on top.

\begin{table}[t]
\centering
\footnotesize
\setlength{\tabcolsep}{3pt}
\begin{tabular}{@{}l rrr r rrr r@{}}
\toprule
& \multicolumn{4}{c}{1080p} & \multicolumn{4}{c}{4K} \\
\cmidrule(lr){2-5} \cmidrule(lr){6-9}
Scene & gsplat & AccuTile & Ours & Reduction & gsplat & AccuTile & Ours & Reduction \\
\midrule
bicycle & 12.0M &  7.56M & 2.15M & 82\% & 31.9M & 15.98M & 2.95M & 91\% \\
bonsai  &  6.6M &  4.16M & 0.74M & 89\% & 21.2M & 11.98M & 1.21M & 94\% \\
counter & 10.6M &  6.39M & 1.05M & 90\% & 34.3M & 18.22M & 1.80M & 95\% \\
garden  & 11.9M &  8.22M & 2.60M & 78\% & 29.5M & 16.99M & 3.41M & 88\% \\
kitchen & 12.0M &  8.06M & 1.81M & 85\% & 34.8M & 20.65M & 2.69M & 92\% \\
room    &  9.1M &  5.31M & 0.77M & 91\% & 30.5M & 16.00M & 1.39M & 95\% \\
stump   &  8.4M &  5.39M & 1.45M & 83\% & 22.9M & 12.16M & 2.03M & 91\% \\
\midrule
\textbf{Mean} & \textbf{10.1M} & \textbf{6.44M} & \textbf{1.51M} & \textbf{85\%} & \textbf{29.3M} & \textbf{16.00M} & \textbf{2.21M} & \textbf{92\%} \\
\bottomrule
\end{tabular}
\caption{Tile--gaussian pair counts (averaged across test cameras, Mip-NeRF~360 scenes, $8\times8$ render tiles). \emph{gsplat}~\cite{ye2024gsplat}: render-tile partitioning with opacity-aware AABB~\cite{radl2024stopthepop}. \emph{AccuTile}: same opacity-aware AABB refined by per-row ellipse intersection~\cite{hanson2024speedysplat}, still at render-tile granularity. \emph{Ours}: the same ellipse-refined intersection at macro-tile granularity. \emph{Reduction} denotes pair count decrease of our scheme with respect to gsplat.}
\label{tab:pairs}
\end{table}

\subsection{Segmented depth sort}
\label{sec:sort}

A segmented sort partitions the input into contiguous segments defined by an offset array and sorts each segment independently.
Macro-tile binning produces exactly this structure: each of the $N_M$ macro-tiles defines one segment containing its binned gaussians, and the sort key is depth.
Segments are independent by construction---no synchronization or data exchange occurs between them.

The segmented formulation also halves the sort key width.
In gsplat~\cite{ye2024gsplat}, a global sort over all render tiles must distinguish which tile each entry belongs to, requiring composite keys that encode both the tile index and the depth---64~bits (32~bits for the tile-encoded index, 32~bits for depth).
Because our sort operates within a single macro-tile, the segment index implicitly encodes spatial location: the key is the 32-bit depth value alone.
Halving the key width halves the radix sort's memory traffic per pass and reduces the number of radix passes from $\lceil 64/r \rceil$ to $\lceil 32/r \rceil$ for radix width $r$.
Sch\"utz~\etal~\cite{schutz2025splatshop} observed the benefit of narrower sort keys within the same factorization, decomposing the 48-bit composite into a 32-bit depth sort of visible splats followed by a 16-bit tile-ID sort of the resulting fragments.
The segmented formulation goes further: because each macro-tile defines an independent sort boundary, the tile-ID dimension is unnecessary and the sort operates on fewer entries---macro-tile pairs rather than render-tile fragments.

Beyond key-width savings, the total entry count drops from the render-tile pair count to $N_\text{macro}$ macro-tile pairs (\cref{tab:pairs}), and segment-level parallelism replaces the single global sort with $N_M$ independent in-segment sorts.
As resolution increases, the render-tile pair count grows but per-segment workload stays bounded---more macro-tiles distribute gaussians into more, smaller segments.
The practical consequence is resolution-stable sort performance: at 4K, our cascade sort remains below 0.08~ms across all benchmark scenes versus 2.5--4.2~ms for a global radix sort over the same content (\cref{tab:sort}).

\subsection{Macro-tile rasterization}
\label{sec:fused-raster}

3DGS materializes explicit per-tile gaussian lists---bitmask generation, counting, prefix sum, fill---before rasterization can begin.
Our macro-tile rasterizer replaces these four stages with inline visibility determination during the gaussian load phase (\cref{fig:fused-raster}), at low marginal cost.
Each processing unit consists of $W$ worker groups of $P$ threads each and executes the rasterization work for its (macro-tile, gaussian-batch) pair in two phases followed by a post-blend compositing pass.
Two primitives recur in the pseudocode (Algorithms~\ref{fig:fused-raster} and~\ref{fig:blend-detail}): $\textsc{Compact}(\mathit{mask})$ returns the positions of the set bits of \emph{mask} as a variable-length list of indices (a stream-compaction primitive), and $\textsc{TransposeMasks}$ transposes the $P{\times}M$ bit matrix of a mini-batch, turning per-gaussian tile masks into per-tile gaussian masks.

\begin{algorithm}[t]
\caption{\textsc{MacroTileRasterize}$(mt, \mathit{batch})$. One processing unit handles a (macro-tile, batch) pair. Phase~1 parallelizes across gaussians (one per thread) to hide memory latency; each thread tests its gaussian against all $M$ tiles and \textsc{TransposeMasks} converts per-gaussian masks to per-tile masks. Worker group~0 builds a \emph{tile work queue} $\mathcal{Q}$. In Phase~2, worker groups dynamically claim tiles from $\mathcal{Q}$---no group idles while tiles remain. (see \cref{sec:gpu-mapping} for implementation details.)}
\label{fig:fused-raster}
\SetKwComment{Comment}{\quad$\triangleright$\ }{}
\KwIn{$W$ worker groups of $P$ threads each; local storage $\mathcal{S}$\newline
      $M$ = render tiles per macro-tile; $B_r$ = gaussians in this batch;\newline
      a \emph{mini-batch} = $P$ gaussians (one per thread)}
\BlankLine
$K \gets \lceil B_r / P \rceil$\Comment*[r]{mini-batches per batch}
\tcp{Phase 1: Load + Inline Visibility (all worker groups)}
\For(\Comment*[f]{groups stride over mini-batches}){$k \gets \mathit{group\_id}$ \KwTo $K{-}1$ \textbf{step} $W$}{
    $j \gets k \cdot P + \mathit{thread\_id}$\Comment*[r]{one gaussian per thread}
    \tcp{gather gaussian data and convert to local tile-space frame}
    $\mathcal{S}.\mathit{pos}[j],\,\mathcal{S}.\mathit{conic}[j],\,\mathcal{S}.\mathit{color}[j] \gets \textsc{LoadGaussian}(mt, j)$
    \;\tcp{$M$-bit mask: bit $t \leftrightarrow$ tile $t$}
    $\mathit{tile\_mask} \gets \textsc{OverlapTest}(\mathcal{S}.\mathit{pos}[j],\,\mathcal{S}.\mathit{conic}[j])$
    \;\tcp{transpose mask from per-gaussian $\to$ per-tile}
    $\mathcal{S}.\mathit{masks}[k{\cdot}M + \mathit{thread\_id}] \gets \textsc{TransposeMasks}(\mathit{tile\_mask})$
}
\textsc{Barrier}()\;
\tcp{Transition: Build work queue (worker group 0 only)}
\If{$\mathit{group\_id} = 0$}{
    $\mathit{hit} \gets \bigvee_{k=0}^{K-1} \mathcal{S}.\mathit{masks}[k{\cdot}M + \mathit{thread\_id}]$\Comment*[r]{OR-reduce per tile}
    $\mathit{active} \gets \textsc{Vote}(\mathit{hit} \neq 0) \;\wedge\; \mathit{InBounds}$\Comment*[r]{valid tile mask}
    $\mathcal{Q} \gets \textsc{Compact}(\mathit{active})$\Comment*[r]{tile work queue in local storage}
}
\textsc{Barrier}()\;
\tcp{Phase 2: Tile Rasterization (all worker groups)}
\While(\Comment*[f]{groups pull tiles dynamically}){$t \gets \textsc{Claim}(\mathcal{Q})$}{
    \textsc{RasterizeTile}$(t, \mathcal{S})$\Comment*[r]{$\to$ Algorithm~\ref{fig:blend-detail}}
}
\end{algorithm}

\begin{algorithm}[t]
\caption{\textsc{RasterizeTile}$(t, \mathcal{S})$. One worker group rasterizes one tile claimed from $\mathcal{Q}$ (Algorithm~\ref{fig:fused-raster}, line~15). The scheduling buffer $\mathcal{R}$ (capacity $2B_d$) decouples variable-rate mask compaction (lines~4--5) from fixed-rate blending (lines~6--7): \textsc{FrontToBackBlend} always processes $B_d$ gaussians from local storage via front-to-back alpha compositing. Early-out (line~8) skips remaining mini-batches when all pixels saturate.}
\label{fig:blend-detail}
\SetKwComment{Comment}{\quad$\triangleright$\ }{}
\KwIn{$P$ threads, each owns $\mathit{tile\_size}^2 / (2P)$ pixel-pairs\newline
      $B_d$ = drain batch size; $K$ = mini-batches per batch (Alg.~\ref{fig:fused-raster});\newline
      $\mathcal{R}$ = scheduling buffer}
\BlankLine
$(R, G, B, T) \gets (0, 0, 0, 1)$\Comment*[r]{per-pixel-pair accumulators}
$\mathcal{R} \gets \varnothing$\Comment*[r]{scheduling buffer, capacity $2B_d$}
\For(\Comment*[f]{iterate over mini-batches}){$k = 0$ \KwTo $K{-}1$}{
    $m \gets \mathcal{S}.\mathit{masks}[k{\cdot}M + t]$\Comment*[r]{$P$-bit: bit $g$ = gaussian $g$ overlaps $t$}
    $\mathcal{R}.\textsc{EnqueueAll}(k{\cdot}P + \textsc{Compact}(m))$\Comment*[r]{bulk push gaussian indices}
    \While(\Comment*[f]{drain full batches}){$|\mathcal{R}| \geq B_d$}{
        \textsc{FrontToBackBlend}$(\mathcal{R}.\textsc{Dequeue}(B_d),\,\mathcal{S},\,R,G,B,T)$\;
        \If(\Comment*[f]{all pixels saturated}){$\textsc{Unanimous}(\forall p\!: T[p] < \varepsilon)$}{
            \textbf{goto} line~\ref{line:store}\Comment*[r]{early-out}
        }
    }
}
\textsc{FrontToBackBlend}$(\mathcal{R}.\textsc{Flush}(),\,\mathcal{S},\,R,G,B,T)$\Comment*[r]{drain remaining}
\textsc{StreamStore}$(\mathit{tile\_buf}[t],\,R,G,B,T)$ \label{line:store}\Comment*[r]{write partial RGBT}
\end{algorithm}

\noindent\textbf{Phase~1---Load and inline visibility.}
The processing unit cooperatively loads its batch of $B_r$ gaussians into local storage in $K{=}\lceil B_r/P \rceil$ \emph{mini-batches} of $P$ gaussians, one per thread of a worker group.
During the load, each gaussian is tested against all $M$ render tiles using a conservative overlap test on its opacity-threshold ellipse (details in \cref{sec:raster-kernel}), producing a per-gaussian tile-visibility mask.
Materializing such a mask through main memory at preprocessing time, as in the ASIC-targeted GS-TG pipeline~\cite{jo2025gstg}, was observed to serialize bitmask generation with the sort and negate sort-cost savings on commodity GPUs; computing the mask inline during the cooperative load avoids that round-trip entirely.
Phase~1 naturally produces per-gaussian information (which tiles does this gaussian overlap?), but Phase~2 requires per-tile information (which gaussians does this tile need?).
A lightweight transpose (Algorithm~\ref{fig:fused-raster}, line~6) converts the per-gaussian tile masks into per-tile gaussian masks: for each mini-batch, every render tile receives a $P$-bit mask whose bit $g$ marks whether mini-batch gaussian $g$ overlaps it, so the $K$ such masks concatenated give the tile's full visibility over the $B_r$ batch gaussians.
This replaces the standard four-stage render-tile refinement pipeline with a single pass.

\noindent\textbf{Phase~2---Per-tile rasterization.}
Active render tiles are assigned to worker groups via the \emph{tile work queue} $\mathcal{Q}$ (Algorithm~\ref{fig:fused-raster}, lines~15--16).
For each claimed tile, the worker group iterates over the precomputed mask, extracting only the visible gaussians from local storage.
A \emph{scheduling buffer} $\mathcal{R}$ accumulates gaussian indices from the mask (Algorithm~\ref{fig:blend-detail}, lines~4--5), then drains them in fixed-size drain batches of $B_d$ gaussians through an unrolled blending core (Algorithm~\ref{fig:blend-detail}, line~7).
This decouples variable-rate mask iteration---some tiles have many visible gaussians, others few---from the fixed-rate blending arithmetic, keeping utilization high regardless of per-tile density.
Blending uses front-to-back alpha compositing with half-precision accumulators tracking per-pixel RGBT.
After each drain batch, a unanimous vote tests whether all pixels in the tile have reached transmittance saturation (Algorithm~\ref{fig:blend-detail}, line~8); if so, the worker group immediately claims the next tile from the queue.

\noindent\textbf{Post-rasterization compositing.}
Because each macro-tile's depth-sorted gaussian list may span multiple batches processed by independent processing units, per-tile results are partial: each unit accumulates color and transmittance only for its batch's visible gaussians.
Correct compositing is possible because front-to-back alpha blending is associative over contiguous depth ranges~\cite{porter1984compositing}.
If batch $i$ produces partial color $C_i$ and residual transmittance $T_i$ for a tile, the final pixel color is
\begin{equation}
    C = C_0 + T_0 \cdot C_1 + T_0 T_1 \cdot C_2 + \cdots,
    \label{eq:compose}
\end{equation}
with early termination when $\prod_i T_i < \epsilon$.
Each processing unit writes its partial RGBT only for tiles that have visible gaussians, using a sparse buffer indexed by a per-batch active-tile bitmask.
A lightweight post-blend pass walks each render tile's batch chain in depth order, accumulating \cref{eq:compose} front-to-back with transmittance early-out.
The pass is sequential across batches per tile but not a bottleneck in practice: each worker group reads a compact stream of partial RGBT records, and tiles that saturate early skip the remaining batches entirely.

The rasterizer realizes the density-proportional decomposition promised in \cref{sec:hierarchy}: processing units scale with gaussian count per macro-tile, render tiles within each unit are distributed dynamically, and the scheduling buffer $\mathcal{R}$ decouples variable-rate filtering from fixed-rate blending.

\subsection{Numerical representation}
\label{sec:numerical}

The rendering pipeline operates entirely in half precision.
The motivation is memory rather than arithmetic: on the commodity GPUs we target, packed half-precision and single-precision share the same math throughput, so the gain is not faster computation but smaller, cheaper data movement.
Storing gaussian attributes in 16~bits halves their footprint in DRAM and in on-chip local storage.
This doubles effective load/store throughput at each level, lets a processing unit keep twice as many gaussians resident, and halves the per-pixel accumulator state each worker carries in the rasterizer's inner loop.
Running the whole pipeline in a single format additionally removes the per-element conversions that a 16-bit-storage, 32-bit-compute design would pay on every load and store.
Two half-precision hazards arise in the per-pixel Mahalanobis distance $q = (\mathbf{p} - \boldsymbol{\mu})^\top \Sigma^{-1} (\mathbf{p} - \boldsymbol{\mu})$, where $\mathbf{p}$ is the pixel position, $\boldsymbol{\mu}$ the projected gaussian center, and $\Sigma$ its $2{\times}2$ screen-space covariance: a \emph{cancellation hazard} in the algebraic form of $q$, and an \emph{overflow hazard} in the magnitudes of the operands that build it.
HiGS addresses each in the data representation rather than at runtime.

Cancellation arises when $\Sigma^{-1}$ is stored as the conic coefficients $(A, B, C)$: the quadratic form $A \cdot dx^2 + 2B \cdot dx \cdot dy + C \cdot dy^2$ in the pixel offset $(dx, dy) = \mathbf{p} - \boldsymbol{\mu}$ has a mixed-sign cross term that is prone to catastrophic cancellation in half precision.
HiGS stores $\Sigma^{-1}$ as its Cholesky factors $\Sigma^{-1} = L L^\top$, so the Mahalanobis distance becomes a sum of squares: $q = (l_0 \cdot dx + l_1 \cdot dy)^2 + (l_2 \cdot dy)^2$.
Each term is the square of a half-precision product and therefore non-negative regardless of rounding error, eliminating the precision collapse that would otherwise force single-precision arithmetic in the inner loop.

Overflow arises when pixel and gaussian coordinates carry their absolute screen-space magnitudes into the products $l_k \cdot dx$ and $l_k \cdot dy$; at high resolution and with anisotropic gaussians, these products can approach the fp16 range limit.
HiGS stores positions relative to the macro-tile center, keeping every coordinate, and every intermediate product entering the Mahalanobis evaluation, well within the fp16 range; the specific units and bounds are deferred to \cref{sec:raster-kernel}.
TC-GS~\cite{liao2025tcgs} identifies the same overflow hazard in its tensor-core polynomial evaluation and addresses it with an equivalent runtime shift (global-to-local coordinates); HiGS makes the shift part of the data layout, so it costs no runtime arithmetic.

Both hazards become properties of the data representation rather than of runtime arithmetic.
Compared against an fp32 reference rasterizer with the gaussian model held fixed, HiGS reaches 67.0~dB PSNR without SH compression and 55.6~dB with the 32-byte SH-compression front end enabled by default (\cref{tab:iq-comparison}); the compositing algorithm itself is exact, and against ground-truth images both configurations produce no observable image-quality loss compared to the gsplat baseline (\cref{sec:quality}).
The mapping of this representation to CUDA implementation is deferred to \cref{sec:raster-kernel}.

\section{Implementation}
\label{sec:impl}

While the architecture of \cref{sec:arch} is hardware-agnostic, this section targets NVIDIA Blackwell GPUs.
The subsections walk \cref{fig:pipeline} in pipeline order: GPU mapping overview, projection and SH evaluation, gaussian partitioning, segmented depth sort, and macro-tile rasterization.

\subsection{GPU mapping overview}
\label{sec:gpu-mapping}

The abstract concepts from \cref{sec:arch} map directly to GPU execution primitives.
For the rasterization kernel, the concrete mapping is:

\begin{itemize}
    \item \emph{Processing unit} = one CUDA CTA (cooperative thread array): $W{=}10$ warps of $P{=}32$ threads (320 threads total).
    \item \emph{Worker group} = one warp ($P{=}32$ threads): each warp corresponds to exactly one render tile.
    \item \emph{Local storage} = shared memory (${\sim}27$~KB per CTA).
    \item \emph{Scheduling buffer} = per-warp ring buffer in shared memory, drain batch $B_d{=}64$.
\end{itemize}

\subsection{Projection and SH evaluation}
\label{sec:fused-proj}

A single kernel replaces three separate DRAM-bound passes (projection, view-direction computation, SH evaluation).
Each thread processes one gaussian end-to-end: camera-space transformation, 2D covariance via Jacobian projection, Cholesky decomposition of the inverse covariance, inline view-direction computation, and degree-3 spherical harmonics evaluation.
The outputs---2D positions, Cholesky conics, colors, and a compact per-warp visibility bitfield---are consumed by the partitioning and rasterization stages.

As an optional bandwidth optimization, SH coefficients can be stored in a 32-byte packed representation.
The compression applies YCoCg-R color-space conversion~\cite{malvar2008lifting} to decorrelate channels, square-root gamma quantization (6~bits luma, 4~bits chroma), and per-basis percentile scaling computed from visible gaussians.
Decoding uses a pure fp16 SIMD pipeline with PTX~\cite{nvidia2025ptx} byte-permute instructions for bitfield extraction, avoiding integer-to-float conversions.
Per-gaussian SH bandwidth drops from 192~bytes (fp32) to 32~bytes---$6\times$ compression at 55~dB mean PSNR (\cref{tab:iq-comparison}).
Inline decoding from this 32-byte representation reduces projection kernel runtime by 23\% on average relative to streaming uncompressed fp16 coefficients.
On large scenes (5--6\,M gaussians), this stage is the dominant frame-time cost of the HiGS pipeline at 37--42\,\% of total frame time (\cref{fig:timing}), since per-gaussian projection and SH evaluation scale with gaussian count regardless of pipeline architecture.

\subsection{Gaussian partitioning}
\label{sec:partitioning}

The count-then-fill binning of \cref{sec:binning} is realized as two CUDA passes over partitioning batches of $B_p{=}8{,}192$ gaussians each.
Both passes consume the per-warp visibility bitmask emitted by the projection front-end (\cref{sec:fused-proj}), so culled gaussians never reach the intersection test.

\noindent\textbf{Count pass.}
One CTA per batch builds a private shared-memory histogram of macro-tile hits using the modified AccuTile intersection~\cite{hanson2024speedysplat} on the Cholesky factors of the inverse covariance.
The histogram is flushed to DRAM as packed uint16 pairs, producing an $N_B \times N_M$ matrix where $N_g$ is the scene's gaussian count, $N_B = \lceil N_g / B_p \rceil$ is the number of partition batches, and $N_M$ is the number of macro-tiles in the frame ($N_M \approx 10^3$ at 1080p, $4 \times 10^3$ at 4K).
With each count stored as a 16-bit unsigned integer (two cells per 32-bit word), the histogram occupies $2 N_B N_M$ bytes---small relative to the gaussian set itself.

\noindent\textbf{Prefix sum.}
A two-level scan over the DRAM histogram widens the packed 16-bit batch counts to 32-bit totals and global write offsets, producing two arrays in a single pass: per-macro-tile total gaussian counts, which become the segment boundaries consumed by the sort (\cref{sec:sort-impl}); and per-batch starting write offsets, one base per (batch, macro-tile) cell, used by the fill pass below.

\noindent\textbf{Fill pass.}
The second pass replays the per-batch intersection using the same visibility bitmask, and each hit deposits a (depth, gaussian-id) record at its precomputed offset.
Depth is stored as raw IEEE~754 float bits reinterpreted as int32, exploiting the property that positive floats sort correctly under unsigned integer comparison.
The output pair list occupies $8 N_\text{macro}$ bytes for $N_\text{macro}$ macro-tile pairs (4-byte depth key + 4-byte gaussian-id; \cref{tab:pairs}).

Together with the sort cascade below, partitioning constitutes the intersection+sort group in \cref{fig:timing}; binning accounts for roughly 60--70\,\% of that group's time (0.10--0.13\,ms at 1080p), with the remainder spent in the sort cascade (\cref{tab:sort}).

\subsection{Segmented depth sort}
\label{sec:sort-impl}

\noindent\textbf{Tier cascade.}
The segmented depth sort (\cref{sec:sort}) is implemented as a four-tier block radix sort cascade followed by a persistent merge phase for overflow segments.
The entire pipeline is fully asynchronous: all kernel launches return immediately, with no host-device synchronization until the sorted results are consumed by the rasterizer.
Each tier sorts the depth keys produced in \cref{sec:partitioning} via four 8-bit radix passes using CUB's block radix ranking primitives~\cite{merrill2010revisiting}.

\begin{table}[h]
\centering
\small
\begin{tabular}{@{}cccc@{}}
\toprule
Tier & CTA Size & Capacity & Role \\
\midrule
T0 & 64  & $\leq$1{,}024 & Majority of segments \\
T1 & 128 & 1{,}025--2{,}048 & \\
T2 & 256 & 2{,}049--4{,}096 & \\
T3 & 512 & 4{,}097--8{,}192 & Emits overflow descriptors \\
\bottomrule
\end{tabular}
\end{table}

Each tier is launched with one CTA per segment; each CTA checks whether its segment falls in the tier's size range and either sorts or early-returns.
Adjacent tiers overlap execution via Programmatic Dependent Launch~\cite{nvidia2025pdl} (PDL, SM~$\geq$~9.0), avoiding a full kernel-completion barrier between independent size classes and hiding tier-launch latency.

\noindent\textbf{Overflow path.}
Segments exceeding 8{,}192 items are handled by an overflow path.
The T3 epilog splits each overflow segment into 8{,}192-item blocks and emits overflow descriptors to device memory.
A persistent block-sort kernel then claims and sorts these blocks via an atomic work counter, with no host synchronization.
The sorted blocks are merged via a binary merge tree constructed device-side: leaf nodes merge pairs of adjacent 8{,}192-item runs, and internal nodes double the run size at each level.
A persistent merge kernel processes this tree through a device-side FIFO work queue---CTAs claim merge partitions via atomic operations, and upon completion enqueue the parent node's partitions when all children are finished.
Each merge partition uses a GPU merge-path algorithm~\cite{green2012gpu}: CTA-level partitioning via warp-cooperative binary search determines balanced sub-problems, which are then merged sequentially in shared memory.

\noindent\textbf{Asynchronous coordination.}
The entire cascade---tiered block sorts, persistent overflow sort, and persistent merge---executes as a sequence of non-blocking kernel launches on a single CUDA stream.
Inter-kernel coordination (overflow descriptors, merge-tree metadata, FIFO queue state) is communicated entirely through device-side atomic counters and memory fences.
Sort is no longer a frame-time bottleneck: it accounts for 30--41\,\% of intersection+sort time and under 0.08\,ms even at 4K (\cref{tab:sort}).

\subsection{Macro-tile rasterization}
\label{sec:raster-kernel}

\noindent\textbf{Numerical format.}
The Cholesky factorization from \cref{sec:numerical} is realized in fp16 through prescaled factors, macro-tile-relative coordinates, and half2 packing.
The factors $L = \bigl[\begin{smallmatrix} l_0 & 0 \\ l_1 & l_2 \end{smallmatrix}\bigr]$ are prescaled to absorb the $\ln \to \log_2$ base conversion and tile-unit coordinate scaling ($l'_k = l_k \cdot \sqrt{\log_2(e)/2} \cdot \text{tile\_size}$), so that the rasterizer evaluates $\alpha = \text{opacity} \cdot \exp_2(-q)$ as a single half-precision $\exp_2$ instruction with no runtime base conversion.
Pixel and gaussian positions are stored in tile units relative to the macro-tile center, with $|x| \le 4$ and $|y| \le 2$ for the $8\times4$ macro-tile layout.
This bound, combined with the prescaled $L$ factors above, keeps every operand entering the Mahalanobis evaluation comfortably below the fp16 range limit (\cref{sec:numerical}).
Throughout the pipeline, half2 packing processes two elements per instruction: the visibility test evaluates two tile columns simultaneously, and the rasterizer blends two pixel rows per accumulator pair.
Three properties motivate the fp16 representation in this kernel.
Shared-memory throughput is 4~bytes per cycle per thread, so packing gaussian attributes as half2 both halves the shared memory they occupy and doubles the rate at which the blending loop reads them.
Native fp16 evaluation avoids the fp16-to-fp32 conversions that a mixed-precision path would incur on every operand.
Finally, fp16 halves register pressure, which matters here because the kernel maps relatively few threads to each tile's pixels, so each thread holds many per-pixel accumulators.

\noindent\textbf{Mask transpose and tile dispatch.}
The per-tile mask transpose (\cref{fig:fused-raster}, line~6) is implemented as a 5-stage warp-shuffle butterfly that transposes the $32\times32$ bit matrix in registers so that lane~$t$ holds a word where bit~$j$ indicates gaussian~$j$ overlaps tile~$t$---no atomics and no shared-memory bank conflicts.
The tile work queue (\cref{fig:fused-raster}, lines~9--13) is built by warp~0 against a single shared-memory counter, keeping the contention warp-local.

\noindent\textbf{Blending loop and post-rasterization compositing.}
The scheduling buffer (\cref{fig:blend-detail}, line~2) is a $2B_d$-entry ring buffer in shared memory.
Gaussian indices are drained in batches of $B_d$, with the blending loop $8\times$ unrolled and indices loaded as batched 16-byte transactions (8 uint16 per load).
The blending inner loop executes ${\sim}$11 half2 operations per pixel-pair per gaussian: the position delta $dx$ and base Cholesky term $u_0^\text{base} = l_0 \cdot dx$ are hoisted outside the pixel-pair loop so that only the $dy$-dependent terms run per pair, and each gaussian requires just 4 shared-memory loads regardless of how many pixel-pairs the thread owns.
After each drain batch, a warp-unanimous ballot vote tests whether all pixel-pairs have saturated transmittance (\cref{fig:blend-detail}, line~8); if so, the warp immediately claims the next tile.
Partial RGBT results are written to a global-memory tile buffer via cache-bypassing streaming stores, so the partial buffers do not evict reused gaussian data from cache.
The buffer itself is preallocated at frame start as a dense array indexed by (macro-tile batch, render tile within macro-tile), occupying $16$\,KB per batch slot at the default $8 \times 8$ render-tile size ($32$ tiles per macro-tile $\times$ $64$ pixels per tile $\times$ 4 RGBT channels $\times$ 2 bytes for half precision); the total footprint is $16 N_R$\,KB for $N_R$ macro-tile gaussian batches, and is the explicit cost of the per-batch parallelism that lets independent processing units write without coordination.
Sparsity is recorded out-of-band: a per-batch 32-bit \emph{active mask} flags which render tiles received gaussians, and the post-blend kernel reads only the flagged slots, leaving the dense buffer's unused entries untouched.

A separate compositing kernel assigns one warp per render tile, walking the parent macro-tile's batches in depth order and reading partial RGBT from the sparse tile buffer only when the per-batch active-tile bitmask has the tile's bit set.
The accumulation follows \cref{eq:compose} front-to-back, with transmittance early-out skipping remaining batches once all pixels saturate.
On small scenes (1.2--1.9\,M gaussians), the macro-tile rasterizer is the dominant cost at 48--51\,\% of HiGS frame time (\cref{fig:timing}); the post-blend pass itself contributes a small fraction since most tiles saturate before the trailing batches arrive.

\begin{table}[t]
\centering
\small
\setlength{\tabcolsep}{2.5pt}
\begin{tabular}{@{}l r rrrr rrrr@{}}
\toprule
& & \multicolumn{4}{c}{1080p (ms)} & \multicolumn{4}{c}{4K (ms)} \\
\cmidrule(lr){3-6} \cmidrule(lr){7-10}
& & \multicolumn{2}{c}{$8\times8$} & \multicolumn{2}{c}{$16\times16$} & \multicolumn{2}{c}{$8\times8$} & \multicolumn{2}{c}{$16\times16$} \\
\cmidrule(lr){3-4} \cmidrule(lr){5-6} \cmidrule(lr){7-8} \cmidrule(lr){9-10}
Scene & Gaussian\# & gsplat & Ours & gsplat & Ours & gsplat & Ours & gsplat & Ours \\
\midrule
bicycle & 6.1M & 3.51 & \textbf{0.76} & 2.81 & 0.89 &  8.13 & \textbf{1.05} & 4.88 & 1.16 \\
bonsai  & 1.2M & 1.41 & \textbf{0.29} & 1.02 & 0.35 &  4.94 & 0.57 & 2.22 & \textbf{0.56} \\
counter & 1.2M & 2.22 & \textbf{0.36} & 1.38 & 0.43 &  7.36 & \textbf{0.68} & 3.35 & 0.72 \\
garden  & 5.8M & 3.46 & \textbf{0.83} & 2.77 & 0.99 &  7.54 & \textbf{1.10} & 4.72 & 1.25 \\
kitchen & 1.9M & 2.66 & \textbf{0.49} & 1.80 & 0.59 &  7.90 & \textbf{0.82} & 4.07 & 0.90 \\
room    & 1.6M & 1.94 & \textbf{0.34} & 1.28 & 0.41 &  6.75 & 0.66 & 3.05 & \textbf{0.66} \\
stump   & 5.0M & 2.24 & \textbf{0.55} & 1.87 & 0.62 &  5.84 & 0.87 & 3.20 & \textbf{0.84} \\
\midrule
\textbf{Mean} & --- & 2.49 & \textbf{0.52} & 1.85 & 0.61 & 6.92 & \textbf{0.82} & 3.64 & 0.87 \\
\bottomrule
\end{tabular}
\caption{Total frame time (ms, median). Bold marks each pipeline's best tile size. At matched $8\times8$: \textbf{4.8$\times$} at 1080p, \textbf{8.4$\times$} at 4K. Tile-agnostic (best vs best): \textbf{3.6$\times$} at 1080p, \textbf{4.4$\times$} at 4K.}
\label{tab:perf}
\end{table}

\section{Results}
\label{sec:results}

\subsection{Experimental setup}
\label{sec:setup}

We evaluate on seven scenes from the Mip-NeRF~360 dataset~\cite{barron2022mipnerf360}, spanning outdoor (bicycle, garden, stump) and indoor (bonsai, counter, kitchen, room) environments with gaussian counts ranging from 1.2M to 6.1M (\cref{tab:perf}), and on the nvcampus park capture for scaling experiments at six gaussian-budget caps from 5M to 75M (\cref{fig:park-perf-scaling}).
All comparisons use the publicly-released Inria 3DGS pre-trained models (30K iterations, SH degree~3), the same source used by every prior-work rasterizer in \cref{tab:fps-comparison}.

All timings are 100-iteration medians across the full COLMAP test camera split per scene (16--39 cameras depending on scene).
The \emph{baseline} throughout is the unmodified gsplat renderer implementing the standard tile-based pipeline, taken at commit~\texttt{ed2557f}; all other comparisons are against published prior work.
Hardware is an NVIDIA RTX PRO 6000 Blackwell (SM~120, 188~SMs).
The default tile size is $8\times8$; we also evaluate $16\times16$ for tile-agnostic comparison. 32-byte SH compression is enabled for our pipeline.

\subsection{Performance}
\label{sec:results-perf}

\begin{figure}[t]
\centering
\includegraphics[width=0.75\linewidth]{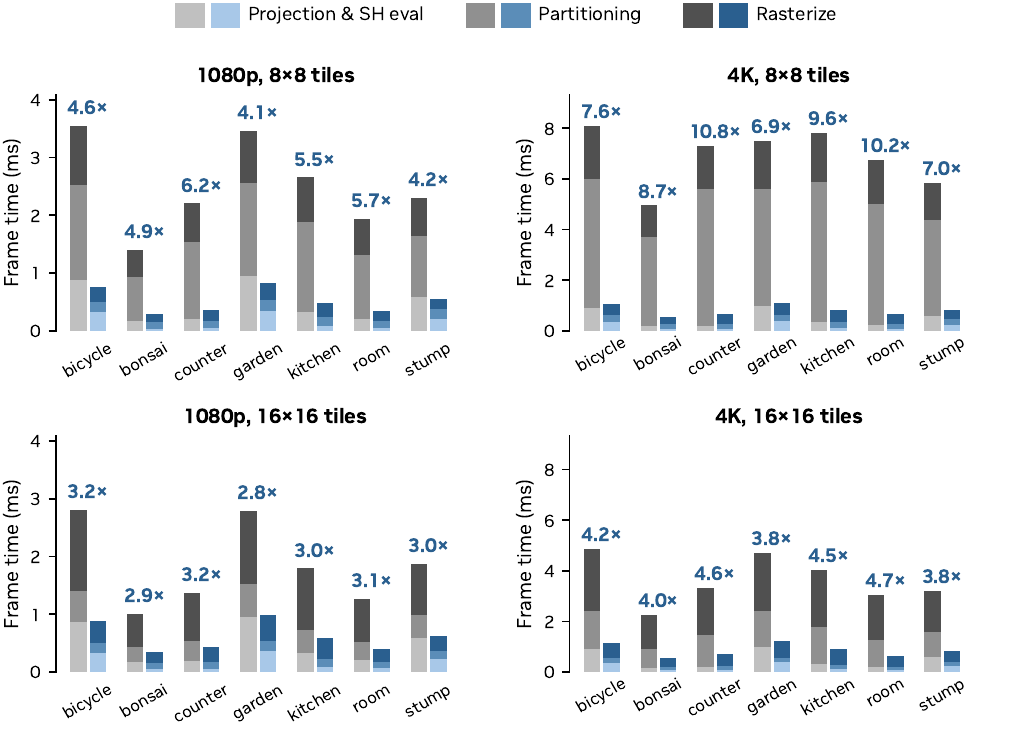}
\caption{\textbf{Per-stage timing breakdown} across seven Mip-NeRF~360 scenes. Top: $8\times8$ tiles, bottom: $16\times16$ tiles. Gray~=~baseline, blue~=~ours. Each bar stacks projection + SH evaluation (light), partitioning (medium), and rasterization (dark). At $8\times8$ tiles, the baseline is partitioning-dominated; at $16\times16$, the bottleneck shifts to rasterization---but our bars remain nearly unchanged in both cases.}
\label{fig:timing}
\end{figure}

\noindent\textbf{Throughput comparison.}
\Cref{tab:fps-comparison} compares HiGS against six prior-work rasterizers on the same hardware.
HiGS leads every scheme at both 1080p and 4K, with the speedup magnitude reflecting each scheme's overall optimization scope.
The original 3DGS introduces the tile-based pipeline with a fixed $3\sigma$ tile margin, producing the largest gap ($6.8$--$11.9\,\times$).
gsplat, Speedy-Splat, and StopThePop use opacity-aware tile bounds in place of the fixed margin and cluster at $3.0$--$4.4\,\times$.
Faster-GS and FlashGS layer per-pixel rendering-stage optimizations on top of opacity-aware tile bounds, giving the smallest gap ($1.8$--$2.2\,\times$).

\noindent\textbf{Per-stage breakdown.}
The per-stage breakdown (\cref{fig:timing}) explains the magnitude of the gap.
The baseline (gsplat) splits roughly evenly between partitioning (46--60\% of frame time at 1080p) and rasterization (26--33\%) at $8\times8$ tiles; at 4K, partitioning rises to 61--73\% because render-tile pair counts roughly triple from 1080p to 4K (\cref{tab:pairs}), and the global radix sort alone consumes 29--46\% of total baseline frame time at 1080p and 44--57\% at 4K.
HiGS's macro-tile pair count grows only $1.3$--$1.8\,\times$ from 1080p to 4K, against $2.5$--$3.4\,\times$ for every render-tile scheme; with the partition bottleneck cut, HiGS's per-stage profile is preprocessing-dominated on large scenes (37--42\% on bicycle, garden, stump at 5.0--6.1\,M gaussians) and rasterization-dominated on small ones (48--51\% on bonsai, counter, kitchen, room at 1.2--1.9\,M).
The result is resolution-stable frame time: mean 0.52~$\to$~0.82~ms ($1.6\,\times$) from 1080p to 4K against the baseline's 2.49~$\to$~6.92~ms ($2.8\,\times$) on the same scenes (\cref{tab:perf}).
This resolution stability is the quantitative signature of the density-proportional schedule (\cref{sec:hierarchy}): per-unit work is bounded rather than screen-area-proportional (\cref{fig:work-decomp}), so the slowest tile no longer gates the frame as it does under one-block-per-tile dispatch.
This is the rasterizer tail effect that the macro-tile schedule removes by construction.
Within this profile, the 32-byte SH representation (\cref{sec:fused-proj}) cuts the projection kernel's runtime by 23\%, which reduces total frame time by 7\% at 1080p and 4\% at 4K on average across the seven scenes; the larger 1080p figure reflects projection's higher share of frame time at lower resolution.

\noindent\textbf{Tile size sensitivity.}
The macro-tile decomposition is robust to render-tile choice (\cref{tab:perf}).
Larger tiles ($16\times16$) help the baseline by 17--38\% at 1080p and 37--55\% at 4K---the partition cost drops, but rasterization waste grows---and its bottleneck only shifts from partition-dominated to rasterize-dominated.
HiGS is 13--22\% slower at $16\times16$ at 1080p and ranges from $-4$\% to $+14$\% at 4K, because the partition saving was already small.
The fairest comparison is each pipeline at its best tile size: HiGS at $8\times8$ versus the baseline at $16\times16$ yields $3.6\,\times$ at 1080p and $4.4\,\times$ at 4K, matching the headline speedup against gsplat in \cref{tab:fps-comparison}.

\noindent\textbf{Sort cascade.}
\Cref{tab:sort} isolates the depth-sort stage.
At 4K, a global radix sort over 64-bit composite keys reaches 2.5--4.2~ms; a CUB segmented radix sort over our 32-bit macro-tile segments reaches 0.21--0.30~ms; the three-tier custom cascade described in \cref{sec:sort-impl} stays below 0.08~ms regardless of scene---a $42$--$73\,\times$ advantage over the global sort.
Within the macro-tile intersection stage, sort accounts for 30--41\% of the intersection time (0.04--0.07~ms at 1080p); the remainder is the binning count-and-fill passes.
Sort is effectively eliminated as a pipeline bottleneck.

\begin{table}[t]
\centering
\small
\setlength{\tabcolsep}{2.5pt}
\begin{tabular}{@{}l rr rr rr rr@{}}
\toprule
& \multicolumn{2}{c}{Pairs} & \multicolumn{2}{c}{CUB Global Sort} & \multicolumn{2}{c}{CUB Seg. Sort} & \multicolumn{2}{c}{Ours} \\
\cmidrule(lr){2-3} \cmidrule(lr){4-5} \cmidrule(lr){6-7} \cmidrule(lr){8-9}
& 64b & 32b & \multicolumn{2}{c}{(ms)} & \multicolumn{2}{c}{(ms)} & \multicolumn{2}{c}{(ms)} \\
Scene & (ref) & (ours) & 1080p & 4K & 1080p & 4K & 1080p & 4K \\
\midrule
bicycle & 12.0M & 2.15M & 1.19 & 3.92 & 0.15 & 0.28 & 0.07 & 0.07 \\
bonsai  &  6.6M & 0.74M & 0.47 & 2.53 & 0.10 & 0.21 & 0.04 & 0.05 \\
counter & 10.6M & 1.05M & 1.03 & 4.17 & 0.11 & 0.25 & 0.06 & 0.06 \\
garden  & 11.9M & 2.60M & 1.16 & 3.56 & 0.12 & 0.30 & 0.07 & 0.07 \\
kitchen & 12.0M & 1.81M & 1.16 & 4.24 & 0.11 & 0.29 & 0.06 & 0.06 \\
room    &  9.1M & 0.77M & 0.80 & 3.65 & 0.10 & 0.27 & 0.04 & 0.05 \\
stump   &  8.4M & 1.45M & 0.66 & 2.55 & 0.12 & 0.28 & 0.06 & 0.06 \\
\bottomrule
\end{tabular}
\caption{Sort-stage comparison (ms, median). Global CUB sorts 64-bit keys over all render-tile pairs; CUB Segmented and Ours sort 32-bit depth keys over macro-tile pairs. Pair counts are averages at 1080p.}
\label{tab:sort}
\end{table}

\begin{figure}[t]
\centering
\includegraphics[width=\linewidth]{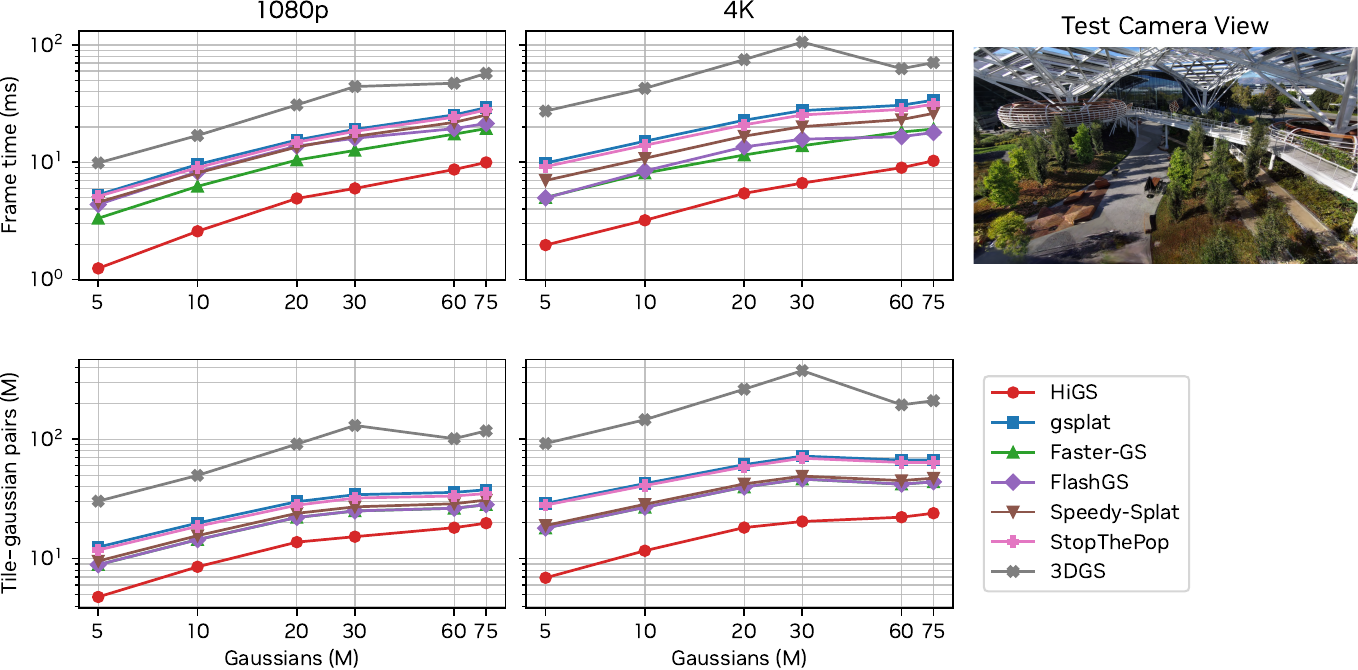}
\caption{\textbf{Cross-scheme scaling on the \emph{nvcampus} park scene} reconstructed at six gaussian-budget caps (5M--75M). \emph{Top row:} median frame time (ms, log scale) vs scene complexity at 1080p and 4K. \emph{Bottom row:} corresponding tile--gaussian pair counts (millions, log scale). The test camera view is shown on the right.}
\label{fig:park-perf-scaling}
\end{figure}

\begin{table*}[t]
\centering
\scriptsize
\setlength{\tabcolsep}{2.5pt}
\begin{tabular}{@{}l r rr rr rr rr rr rr rr rr@{}}
\toprule
& & \multicolumn{2}{c}{Ours (FPS)} & \multicolumn{2}{c}{FlashGS} & \multicolumn{2}{c}{Faster-GS} & \multicolumn{2}{c}{TC-GS} & \multicolumn{2}{c}{Speedy-Splat} & \multicolumn{2}{c}{StopThePop} & \multicolumn{2}{c}{gsplat} & \multicolumn{2}{c}{3DGS} \\
\cmidrule(lr){3-4} \cmidrule(lr){5-6} \cmidrule(lr){7-8} \cmidrule(lr){9-10} \cmidrule(lr){11-12} \cmidrule(lr){13-14} \cmidrule(lr){15-16} \cmidrule(lr){17-18}
Scene & \#Gauss & 1080p & 4K & 1080p & 4K & 1080p & 4K & 1080p & 4K & 1080p & 4K & 1080p & 4K & 1080p & 4K & 1080p & 4K \\
\midrule
bicycle & 6.1M & \textbf{1316} & \textbf{ 949} &  582 & 508 &  599 & 454 &  468 & 356 &  415 & 296 &  374 & 218 &  356 & 205 & 163 &  60 \\
bonsai  & 1.2M & \textbf{3460} & \textbf{1745} & 1543 & 1121 & 1471 & 725 & 1443 & 839 & 1156 & 608 & 1036 & 481 &  981 & 448 & 557 & 167 \\
counter & 1.2M & \textbf{2809} & \textbf{1471} & 1212 & 746 & 1302 & 754 & 1242 & 643 &  925 & 456 &  789 & 317 &  725 & 298 & 343 & 110 \\
garden  & 5.8M & \textbf{1199} & \textbf{ 907} &  591 & 480 &  559 & 426 &  466 & 336 &  418 & 278 &  375 & 225 &  360 & 212 & 222 &  94 \\
kitchen & 1.9M & \textbf{2049} & \textbf{1214} &  852 & 541 &  931 & 568 &  916 & 488 &  679 & 343 &  615 & 265 &  554 & 246 & 295 & 106 \\
room    & 1.6M & \textbf{2967} & \textbf{1515} & 1466 & 907 & 1353 & 773 & 1232 & 708 &  934 & 497 &  809 & 348 &  782 & 328 & 331 & 104 \\
stump   & 5.0M & \textbf{1818} & \textbf{1147} &  912 & 796 &  921 & 626 &  682 & 492 &  625 & 407 &  560 & 341 &  528 & 304 & 371 & 140 \\
\midrule
\textbf{Mean} & --- & \textbf{1937} & \textbf{1214} & 893 & 670 & 897 & 588 & 765 & 499 & 643 & 385 & 573 & 293 & 541 & 275 & 286 & 102 \\
\textbf{Speedup} & --- & --- & --- & 2.17 & 1.81 & 2.16 & 2.06 & 2.53 & 2.43 & 3.01 & 3.16 & 3.38 & 4.14 & 3.58 & 4.42 & 6.77 & 11.86 \\
\bottomrule
\end{tabular}
\caption{Throughput at 1080p and 4K (FPS, $1000/\text{median ms}$)~\cite{feng2024flashgs,hahlbohm2026fastergs,hanson2024speedysplat,liao2025tcgs,radl2024stopthepop,ye2024gsplat,kerbl2023gaussian}. Ours uses $8\times8$ tiles with 32-byte SH compression. Each prior-work scheme runs at its fastest configuration at $16\times16$ tiles with the same single tile-level depth ordering as ours (no per-pixel sort queues, no antialiasing). TC-GS reference works on top of Speedy-Splat rasterizer.}
\label{tab:fps-comparison}
\end{table*}

\noindent\textbf{Scaling with gaussian count.}
\Cref{fig:park-perf-scaling} extends the comparison to the nvcampus park capture across six gaussian-budget caps from 5\,M to 75\,M.
HiGS leads every scheme at every budget at both 1080p and 4K, and its frame time grows roughly linearly in gaussian count---from 1.25 to 9.97\,ms at 1080p and 1.97 to 10.29\,ms at 4K---preserving the resolution-stability property of \cref{tab:perf}.
The gap against partitioning-bound schemes compresses with scene size: at 1080p the ratio over gsplat drops from $4.2\,\times$ at 5\,M to $2.9\,\times$ at 75\,M as per-frame fixed overheads amortize, while 3DGS's 4K frame time grows steeply across the same range (9.85 to 70.67\,ms).
FlashGS and Faster-GS keep a roughly $2\,\times$ gap across the full budget range, mirroring the Mip-NeRF~360 finding above.

\subsection{Image quality}
\label{sec:quality}

\begin{table}[t]
\centering
\small
\setlength{\tabcolsep}{4pt}
\begin{tabular}{@{}l rrr rrr@{}}
\toprule
 & \multicolumn{3}{c}{vs.\ gsplat baseline} & \multicolumn{3}{c}{vs.\ ground-truth} \\
\cmidrule(lr){2-4} \cmidrule(lr){5-7}
Method & PSNR & SSIM & LPIPS & PSNR & SSIM & LPIPS \\
\midrule
Ours w/o SH comp. & 67.03 & .9999 & .0000 & 27.68 & .8649 & .1034 \\
Ours w/ SH comp.  & 55.59 & .9995 & .0007 & 27.67 & .8645 & .1034 \\
FlashGS           & 49.74 & .9983 & .0010 & 27.65 & .8633 & .1046 \\
Faster-GS         & 73.86 & 1.0000 & .0000 & 27.68 & .8649 & .1035 \\
Speedy-Splat      & 94.43 & 1.0000 & .0000 & 27.68 & .8649 & .1034 \\
StopThePop        & 94.37 & 1.0000 & .0000 & 27.68 & .8649 & .1034 \\
3DGS              & 75.04 & 1.0000 & .0000 & 27.68 & .8649 & .1034 \\
gsplat            & ---   & ---   & ---   & 27.68 & .8649 & .1034 \\
\bottomrule
\end{tabular}
\caption{Image quality means under two comparison protocols. \emph{Left:} rendered output vs.\ fp32 gsplat baseline at $1920\times1080$ --- isolates rendering-pipeline differences with the gaussian model held fixed. \emph{Right:} rendered output vs.\ COLMAP test images at factor-4 native resolution --- measures absolute image quality. Metrics: PSNR (dB), SSIM~\cite{wang2004ssim}, and LPIPS~\cite{zhang2018lpips} via AlexNet. The left (vs-gsplat) PSNR is a numerical-precision microbenchmark; against ground truth (right) all schemes lie within 0.04~dB of each other.}
\label{tab:iq-comparison}
\end{table}

HiGS achieves high image quality both against ground truth and against the fp32 reference renderer, despite running fp16 arithmetic throughout the rasterization pipeline (\cref{tab:iq-comparison}).
Against the COLMAP test set HiGS attains 27.68~dB PSNR---the same level as existing fp32 rasterizers---and 67~dB against the gsplat baseline reconstructed output. At 8~bits per channel, 67~dB corresponds to RMS error near $0.1$ of one code value, well below a single quantization step, so the deviation is not representable in the displayed 8-bit image.
The absolute-quality spread across every scheme in the table is just 0.04~dB: rendering-kernel differences all sit below the trained gaussian model's residual error against the test set.

The rendering-kernel comparison against gsplat fp32 baseline spans 45~dB across schemes, and each scheme's offset traces to a specific design choice.
Speedy-Splat and StopThePop reach the ${\sim}94$~dB ceiling, reproducing the fp32 reference to numerical precision.
3DGS sits ${\sim}20$~dB lower because, as noted by StopThePop~\cite{radl2024stopthepop}, its fixed $3\sigma$ tile-touch bound can be too small for the renderer's $1/255$ contribution threshold on sufficiently opaque gaussians, causing above-threshold tile contributions to be omitted.
Faster-GS sits at a similar level for a different reason: it evaluates the transmittance termination threshold after compositing rather than before, and does not apply the $\min(0.99,\,\alpha)$ stability clamp the reference applies during blending.
HiGS's 67~dB level is set by fp16 accumulation and a transmittance termination check evaluated at drain-batch boundaries (every $B_d$ gaussians) rather than per-gaussian; the SH-compressed variant pays an additional 11~dB.
FlashGS reaches 50~dB because its final pixel buffer is 8-bit per channel; internal precision is bounded below the requantization noise floor regardless of upstream choices.

The kernel-level spread is therefore a sensitive microbench rather than a perceptual signal: HiGS trades ${\sim}27$~dB of fp32-reference fidelity for fp16 throughput, drain-batch transmittance, and $6\,\times$ SH bandwidth reduction, and these choices remain imperceptible against ground truth.

\section{Discussion and Limitations}
\label{sec:discussion}

\noindent\textbf{Dataset scope}
Results span seven Mip-NeRF~360~\cite{barron2022mipnerf360} scenes (1.2--6.1\,M gaussians) and the larger nvcampus capture (up to ${\sim}75$\,M gaussians), covering both the standard small-scene benchmark and the multi-tens-of-millions regime relevant to production capture.
Within this range the macro-tile schedule's load-balancing property holds without scene-specific tuning, and the pipeline's per-stage costs (\cref{tab:perf}) extrapolate roughly linearly in gaussian count; we have not validated on scenes outside this scale.
A separate dimension we do not exercise is feature-parity comparison with per-pixel sort queues or antialiasing enabled across schemes.

\noindent\textbf{Forward-only}
The current implementation covers forward rendering; differentiable training is not addressed in this paper.
Most of the architecture---macro-tile decomposition, per-macro-tile sort segments, batched rasterization with sparse active masks---carries over to a backward pass essentially unchanged, but three pieces of bookkeeping change shape.
First, the post-blend reduction that composites partial RGBT slots into a single per-pixel value (\cref{sec:raster-kernel}) becomes a prefix scan: each batch's backward step needs the cumulative RGB and T \emph{before} that batch as starting state, which the forward pass currently produces and discards.
Second, gradient accumulation needs to be re-introduced in fp32 even though forward keeps the per-pixel state in fp16; the algebraic stability of the Cholesky form (\cref{sec:numerical}) carries over, but the division by $(1-\alpha)$ that reverses the alpha-blend amplifies fp16 rounding by up to $\alpha^{-2}$ near saturation and is the natural place for selective upcasting.
Third, the per-pixel last-contributing-gaussian index that backward uses for early termination~\cite{kerbl2023gaussian} becomes a per-(pixel, batch) index in our scheme, since transmittance can saturate inside any batch rather than at a single ordered position.
None of these requires changing the macro-tile decomposition itself, and we expect the extension to be additive.

\noindent\textbf{Hardware specificity}
The architecture itself is hardware-agnostic (\cref{sec:arch}), and its numerical design rests on packed half-precision (half2) storage and arithmetic, available on NVIDIA desktop GPUs since Volta and Turing.
The rasterizer's per-block shared-memory footprint (${\sim}27$~KB; \cref{sec:gpu-mapping}) sits well within the ${\sim}100$~KB per-SM shared memory of Ampere-class and newer desktop SKUs, so we expect the core pipeline to run well across commodity GPUs in that range, though we have not yet characterized performance on Ampere or Ada.
The one launch-time dependency on recent hardware is the PDL-coordinated kernel chain (\cref{sec:sort-impl}), which uses Hopper-class programmatic dependent launch; on architectures without it the same schedule can be expressed with conventional launches at some coordination cost.

\noindent\textbf{Complementary techniques}
Several published optimizations operate at orthogonal pipeline stages: training-time model compression~\cite{fan2024lightgaussian}, Faster-GS's backward-pass acceleration~\cite{hahlbohm2026fastergs}, and SplatShop's interactive editing~\cite{schutz2025splatshop} apply before, after, or alongside our rendering kernel, and large-scale and distributed schemes~\cite{li2024retinags,kerbl2024hierarchical} compose with renderer-side acceleration on a different scaling axis.
FlashGS's~\cite{feng2024flashgs} software-pipelined fetch loop could be adopted in our per-batch rasterizer to overlap memory and compute; tensor-core blending~\cite{liao2025tcgs} would primarily help in a math-bound regime, which is not where our profile sits.
Per-pixel depth-ordering~\cite{radl2024stopthepop} is not directly composable: it would require materializing the per-pixel sort queues our macro-tile scheme deliberately avoids.

\section{Conclusion}
\label{sec:conclusion}

We have presented HiGS, a 3D-gaussian-splatting rendering architecture in which spatial partitioning and rasterization operate at different granularities.
Across seven Mip-NeRF~360 scenes and the nvcampus capture, HiGS leads every prior-work rasterizer we evaluate at both 1080p and 4K, while preserving exact alpha compositing.
We achieve this by decoupling the coarse (macro-tile) spatial partition from the fine (render-tile) rasterization grid, so that segmented sorting, rasterization, cross-tile data reuse, and density-proportional load balancing all follow as architectural consequences of a single decomposition rather than as independent optimizations.

Extending HiGS to differentiable training is a natural next step; the macro-tile decomposition carries over to the backward pass, with the additional per-batch bookkeeping discussed in \cref{sec:discussion}.

\section*{Acknowledgement}
We would like to thank Zan Gojcic for his valuable guidance and support. 
We also thank Jorge Condor Lacambra for providing \emph{nvcampus} dataset.
We are grateful to Ruilong Li, Jean-Eudes Marvie, Qi Wu, and Nicolas Moenne-Loccoz for insightful discussions, feedback, and encouragement.
Finally, we would like to thank David Lesage, Vincent Caux-Brisebois and Eric Shangguan for their help in integrating HiGS into gsplat.

{\small
\bibliographystyle{abbrv}
\bibliography{main}
}
\end{document}